\documentclass{article}

\usepackage[numbers]{natbib}

\usepackage[preprint]{neurips_2021}

\usepackage[utf8]{inputenc} 
\usepackage[T1]{fontenc}    
\usepackage{hyperref}       
\usepackage{url}            
\usepackage{booktabs}       
\usepackage{amsfonts}       
\usepackage{nicefrac}       
\usepackage{microtype}      
\usepackage{xcolor}         

\usepackage{multirow}
\usepackage{graphicx}
\usepackage{fixltx2e}
\usepackage{floatrow}
\usepackage{caption}
\usepackage{wrapfig}
\usepackage{color,soul}

\usepackage{amsmath}
\DeclareMathOperator*{\argmax}{arg\,max}
\usepackage{comment}

\newcommand{\ubold}[1]{\fontseries{b}\selectfont#1}

\title{Multi-Level Contrastive Learning for Few-Shot Problems}

\author{%
  Qing Chen 
  \\
  Department of Computer Science\\
  Louisiana State University\\
  Baton Rouge, LA 70803, USA \\
  \texttt{qchen11@lsu.edu} \\
  \And
  Jian Zhang \\
  Department of Computer Science\\
  Louisiana State University\\
  Baton Rouge, LA 70803, USA \\
  \texttt{zhang@csc.lsu.edu} \\
}

\begin{document}

\maketitle

\begin{abstract}
Contrastive learning is a discriminative approach that aims at grouping similar samples closer and diverse samples far from each other. It it an efficient technique to train an encoder generating distinguishable and informative representations, and it may even increase the encoder's transferability. Most current applications of contrastive learning benefit only a single representation from the last layer of an encoder.In this paper, we propose a multi-level contrasitive learning approach which applies contrastive losses at different layers of an encoder to learn multiple representations from the encoder. Afterward, an ensemble can be constructed to take advantage of the multiple representations for the downstream tasks. We evaluated the proposed method on few-shot learning problems and conducted experiments using the mini-ImageNet and the tiered-ImageNet datasets. Our model achieved the new state-of-the-art results for both datasets, comparing to previous regular, ensemble, and contrastive learing (single-level) based approaches.
\end{abstract}

\section{INTRODUCTION}
\label{intro}

Contrastive learning (CL) aims to learn representations that group similar samples closer and diverse samples far from each other. In recent years, CL has been shown to be successful in learning discriminative features that improve the performance of a model.   
In particular, many CL-based approaches have been proposed to generate effective image features for computer vision tasks \cite{chen2020improved, khosla2020supervised, chen2020simple, tian2020makes, wang2020understanding}. 

Most current contrastive learning work focused on learning a single representation from an encoding network (outputs from one layer, usually the last, of the encoder). The contrastive loss is applied to this representation during training and the learned representation is used in the downstream tasks. However 
a deep neural network has many layers and the neurons at different layers may react to stimuli of different types. 
For example, in a convolutional neural network that processes images, the lower-level neurons have smaller receptive fields and respond to local features while the higher-level ones respond to larger features. At both the local and the global level, similar images should have similar collection of features (if spatial transformations are ignored) while the collection of features from dissimilar ones can vary. Hence, the principle of contrastive learning should be applicable to both the lower and the higher level representations. It is a natural extension to consider contrastive learning at multiple levels. Furthermore, since the higher-level features are dependent on the lower-level ones (upper neurons get their inputs from the lower ones), one may expect that better learning at the lower level may lead to better higher-level features. Therefore, multi-level contrastive learning could be beneficial even when only the representation from the top layer is concerned.  
The work from \citet{lowe2019putting} also motivated us to investigate a multi-level approach to contrastive learning. It showed that hierarchical (multi-level) predictive coding was able to produce high quality representations. The training process applies predictive-coding losses layer-wise at multiple levels. A final representation at the top layer was employed in the downstream classification and showed highly competitive performance.     

In this work, we ask and investigate two problems: first, can multi-level contrastive learning lead to better representations for downstream tasks? By multi-level contrastive learning, we mean a learning process where the contrastive loss is applied at multiple levels (layers) of the network. 
The losses from the layers can be aggregated and optimized together. Or the training can be conducted layer-by-layer when the memory is too limited to hold the full model. The second problem we investigate is how to take advantage of the multiple representations. Multi-level contrastive learning naturally generates multiple representations from different layers. One may use only the representation from the top layer in the downstream task and utilize multi-level contrastive learning as a way to train a better final representation. Alternatively, one may include multiple representations in the downstream tasks if they can be employed to improve the performance.

\begin{figure*}
\centering
\includegraphics[scale=1.1
]{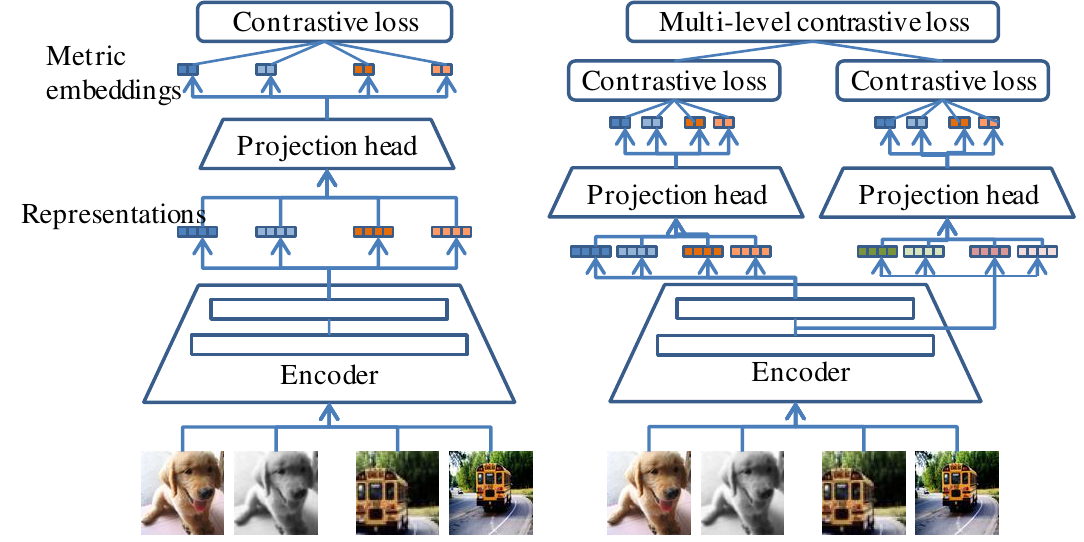}
\caption{Multi-level contrastive learning (right) compared with the standard single-level contrastive learning (left).
}\label{fig:mm}
\end{figure*}

To investigate these problems, we considered the few-shot learning (FSL) problems~\citep{snell2017prototypical} as the testing environment. In FSL, it is critical to train an encoder that produces discriminative features, especially for the unseen classes. This makes it a good scenario for testing.  
Our investigation led us to propose a multi-level contrastive learning approach that extends the standard CL. Specifically, different projection networks are built for the representations at different layers. The contrastive loss is computed at each layer based on the projection. The learning process then optimizes the total loss aggregated from the layers. (See Figure~\ref{fig:mm}.) (In general, we may impose the contrastive losses at individual layer level or we may impose them at the block level, e.g., the blocks used in residual networks.) For the second problem, we adopt an ensemble approach to further take advantage of the multiple representations obtained from the multi-level contrastive learning. Each representation, trained using multi-level contrastive learning, leads to a classifier in the ensemble for the downstream FSL classification. The combination of the multiple-level contrastive learning and the ensemble provides a mechanism to generate and utilize discriminative and informative representations from different levels and therefore, may be able to improve the downstream classification.

Experiments with our proposed model were conducted on two FSL benchmark datasets and we obtained new state-of-the-art results for both of them. 
(Note that in recent years, many FSL works have employed extra data from the test i.e., the novel, classes for better performance. The extra data can be unlabeled and given at the test time, e.g., transductive learning~\citep{kye2020transductive,yang2020dpgn}, or during the training phase, e.g.,  semi-supervised learning~\citep{ rodriguez2020embedding, lichtenstein2020tafssl}. We focus on the traditional FSL setting, where only a few support data points per novel class are available at the test time, and compare to works only in this setting.) 
We also conducted experiments that demonstrated that the utilization of multi-level contrastive learning are beneficial for training an powerful encoder, and multi-representation in the ensemble is crucial for the success in FSL problems.
Our main contributions are as follows:
\begin{enumerate}
\item  We propose a multi-level contrastive learning approach that extends the standard CL and makes the representations from different layers of encoders informative and discriminative.
\item  We adopt an ensemble method that creates a collection of models by employing multiple representations from different depth of a deep neural network. 
\item  We demonstrate the advantage of our approach on the FSL problems and achieved new state-of-the-art results on two benchmark datasets. Our experiments also show that multi-level contrastive learning enhances the efficiency of encoders and the generated multi-representation boost the performance of the ensemble. 
\end{enumerate}

\section{METHODOLOGY}
\label{method}

\subsection{Multi-Level Contrastive Learning}


Multi-level contrastive learning is an extension to the standard self-supervised contrastive learning that applies to the representation at a single level. As illustrated in Figure \ref{fig:mm}, multi-level contrastive learning is composed of the following three major components.

\begin{itemize}
\item A data augmentation module. For an input $x$, the module randomly produces two different augmentations/views of $x$, denoted as $\bar{x}$  and  $\bar{x}'$. Details of the augmentation approach is described in Section \ref{exp}.

\item An encoder network $e(\cdot)$. It generates representation vectors for the augmented examples. We use a convolutional neural network (CNN) here as an example of the encoder but the approach is
not limited to just CNNs. 
We refer to the representations for $\bar{x}$ and $\bar{x}'$ from the $n$-th convolutional layer in the encoder $e$ by $\phi_{e}^{(n)}(\bar{x})$ and $\phi_{e}^{(n)}(\bar{x}')$. 

\item Multiple projection heads $pr_{p}(\cdot)$ where $p$ is one of the selected layers on which contrastive loss is added. $pr_{p}(\cdot)$ projects the representation vectors into a hidden space. Namely, for $\phi_{e}^{(n)}(\bar{x})$ and $\phi_{e}^{(n)}(\bar{x}')$, $pr_{p}(\cdot)$ maps them into vectors $z_p(\bar{x}) = pr_{p}(\phi_{e}^{(n)}(\bar{x}))$ and $z_p(\bar{x}') = pr_{p}(\phi_{e}^{(n)}(\bar{x}'))$. We normalize the vectors produced by $pr_p(\cdot)$ and apply an inner product to measure their distances. 
\end{itemize}

Given the components, we now introduce the details of the multi-level contrastive loss. 
We randomly sample $B$ examples, $x_i$, where $i \in \{1, 2, \ldots, B\}$. After the augmentation, the samples form a batch of size $2B$. The batch is the union $\{\bar{x_i}\} \cup \{\bar{x_{i}}'\}$. 
$\bar{x_i}$ and its correspondence $\bar{x_{i}}'$ are treated as the positive pair while the other $2(B-1)$ examples within the same batch are treated as the negatives. 

Let $i \in I $ where $I = \{1, \dots, 2B\}$ be the index of an arbitrary sample in the batch.
For the sample $\bar{x_i}$, let $j$ be the batch index of its positive pair, i.e., $\bar{x_j} = \bar{x_i}'$. The contrastive loss for the representations from $p$-th layer of an encoder $e$ is similar to the one used in the single-level self-supervised contrastive learning \cite{chen2020simple}, that is

\begin{equation}
L^{p}= \sum_{i \in I} L_i^{p} =  -\sum_{i \in I} \log {{\exp ( \mbox{sim} (z_p(\bar{x_i}),z_p(\bar{x_j}) / \tau) } \over {\sum\limits_{k \in J(i)}\exp ( \mbox{sim} (z_p(\bar{x_i}),z_p(\bar{x_k}) / \tau)}} 
\end{equation}

Here 
sim($u,w$) = $u^\top w / \|{u}\| \|{w}\|$ which is the inner production between $l_2$ normalized $u$ and $w$, $J(i) = I \setminus \{i\}$, and $\tau$ denotes a scalar temperature parameter. The denominator of the equation has a total of $2B -1$ terms with exception of $i$ itself.

For each selected layer on which the contrastive loss is applied, there is a separate projection head $pr(\cdot)$ receiving the representations from that layer. Thus we have multiple projection heads and multiple contrastive losses (See right part of Figure \ref{fig:mm}). We aggregate these losses to get the final multi-level contrastive loss, which is defined as follows

\begin{equation}
L_{MLCL}= \sum_{p \in P} L^{p} = \sum_{p \in P}  \sum_{i \in I} L_i^{p}
\end{equation}

Here, $P$ is the collection of all selected layers. 

\subsection{Taking Advantage of Multiple Representation by an Ensemble} 
\label{mm-training}
\begin{figure*}
\centering
\includegraphics[scale=0.9
]{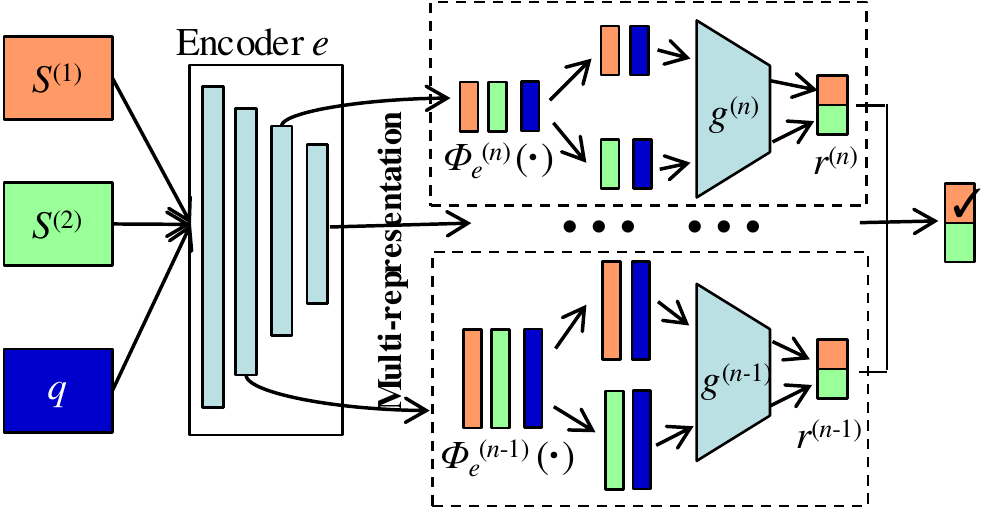}
\caption{Classification by multi-representation ensemble for a 2-way 1-shot problem.}\label{fig:arch}
\end{figure*}

We now introduce the ensemble approach to utilize the multiple representations from the encoder trained using multi-level contrastive learning. The combination of a particular representation and a classifier network forms an individual classification model in the ensemble. 
Figure~\ref{fig:arch} shows the overall architecture of the ensemble. (For simplicity, the figure illustrates only one encoder with multiple representations. An actual ensemble may include several encoders.) In the following, We describe the details of the ensemble.

Our ensemble uses relation network~\citep{sung2018learning} to perform classification. 
The relation network takes a pair of instances from a particular representation as input and outputs a similarity score. We denote by $g$ the score function computed by the relation network. 

Let $S^{(1)}$ and $S^{(2)}$ be two images from two different classes (class 1 and 2) in the support set. Let $q$ be a query image to be classified. The images are first fed into the encoders and for each image, multiple representations (from different encoders and different layers) are generated. 
Take an encoder $e$ and the $n$-th layer of $e$ as an example, We illustrate the classification process using the representations $\phi_{e}^{(n)}(s^{(1)})$ (orange in Figure~\ref{fig:arch}), $\phi_{e}^{(n)}(s^{(2)})$ (green in Figure~\ref{fig:arch}) and $\phi_{e}^{(n)}(q)$ (blue in Figure~\ref{fig:arch}).
(Note that Figure~\ref{fig:arch} contains two sets of representations $\phi_{e}^{(n)}(\cdot)$ and $\phi_{e}^{(n-1)}(\cdot)$ with the same color scheme. Colors are used to indicate the source image from which the representation is computed. They are not related to the location, i.e., the encoder and the layer, where the representation is generated.) The representation of $\phi_{e}^{(n)}(q)$ is concatenated (on the channel dimension) to the representation of $\phi_{e}^{(n)}(s^{(1)})$. The relation network is applied to the concatenation and a score $r$ is calculated to measure the similarity between the query and the class (class 1) representative image $s^{(1)}$, i.e. $r^{(1)}(q) = g(\phi_{e}^{(n)}(q)||\phi_{e}^{(n)}(s^{(1)}))$ where $||$ indicates concatenation on the channel dimension. In an $N$-shot ($N>1$) setting, for each class, we have $N$ examples (e.g.,  $S^{(k)}_1, S^{(k)}_2, \ldots, S^{(k)}_N$ for class $k$). A prototype is obtained by averaging the representations $\phi_{e}^{(n)}(s^{(k)}_i)$. The score is computed using the concatenation of $\phi_{e}^{(n)}(q)$ and the prototype, i.e., in general, 
\begin{equation}
r^{(k)}(q) = g(\phi_{e}^{(n)}(q)||\frac{1}{N}\sum_i\phi_{e}^{(n)}(s^{(k)}_i))
\end{equation}
This calculation is performed for each class
to produce $C$ scores $r^{(1)}, r^{(2)}, \ldots, r^{(C)}$ for a $C$-way classification. The query image is classified into the category that has the maximum score.    

The above presents the classification process of an individual model in the ensemble. There are a set of encoders $e_1, e_2, \ldots$ and for each encoder $e_i$, multiple representations from different layers $n^{(e_i)}_1, n^{(e_i)}_2, \ldots$ are used to produce models in the ensemble. Let $T = \{n^{(e_i)}_j\}$ be the set of representations, each leading to a classifier. Let $r^{(i)}_t(q)$, $t\in T$, $i \in \{1, 2, \ldots, C\}$ be the score for image $q$ with respect to class $i$ and computed by the classifier using representation $t$. The ensemble's final classification is made following the average scores over the models, i.e., the class label $c^*(q)$ is predicted to be:   

\begin{equation}
c^*(q) = \argmax_i \frac{1}{|T|} \sum_{t \in T} r^{(i)}_t(q)
\end{equation}

where $|T|$ is the size of the set $T$. 
Each relation network in the ensemble will be trained independently following the training procedure in~\citep{sung2018learning} while the encoders remain unchanged after pre-training.

\section{EXPERIMENT RESULTS}
\label{exp}


\subsection{Few-Shot Learning Problems}
Formally, we have three disjoint datasets: a training set $D$\textsubscript{train}, a validation set $D$\textsubscript{val}, and a testing set $D$\textsubscript{test}. 
In the traditional $C$-way $N$-shot classification in FSL, we are tasked to obtain a model that can perform classification among $C$ classes (in the testing set $D$\textsubscript{test}) while we have access to only $N$ samples from each class. (The model can be pre-trained using data in $D$\textsubscript{train} or even data in $D$\textsubscript{val}. However, there are no overlapping classes between $D$\textsubscript{test} and $D$\textsubscript{train} or $D$\textsubscript{val}.)  
The set of $C\times N$ samples is often referred to as the support set. 
In many FSL works, $N$ is commonly set to be 1 or 5.  
We remark that some recent FSL researches have started to explore helps from additional information. In particular, a number of methods have been proposed to leverage unlabeled data beyond the $N$ examples from the $C$ classes 
to enhance model accuracy. They either use the unlabeled data in the training (semi-supervised learning) or 
perform classifications with a set of query data together (transductive learning)
\citep{li2019learning,kye2020transductive, yang2020dpgn, hu2020exploiting, rodriguez2020embedding, lichtenstein2020tafssl}. 
In this work, we focus on the traditional FSL setting where no additional information or unlabeled data are available and the prediction for a query data point is made independently from (without knowing) any other query data points. 

\subsection{Experimental Setup}\label{sec:expt_setup}

\textbf{Datasets}. We use two standard benchmark datasets in FSL: mini-ImageNet \citep{vinyals2016matching}  and tiered-ImageNet \citep{ren2018meta}. Both are publicly available. Mini-ImageNet consists of 100 categories, 64 for training, 16 for validation and 20 for testing, with 600 images in each set. Tiered-ImageNet includes 608 classes (779, 165 images) split as 351 training, 97 validation and 160 testing classes, each with about 1300 images.
The image size of both datasets is $84 \times 84$.


\textbf{Implementation details}. We adopt the same ResNet-18 model in \citep{han2020automatically} and two variants of ResNet-18 as the encoders. The first variant is ResNet-18-v1 in \citep{sun2019meta} which has 3 residual sections with 3 basic blocks, each including 2 convolutional layers,  and the second variant is ResNet-18-v2  consisted of 2 residual sections with 4 basic blocks. 

In multi-level contrastive learning framework, we apply data augmentation twice for input examples based on three approaches: random cropping followed by resize back to the original size, random color distortions, and random Gaussian blur. The projection heads are consist of two linear layers, with the first one, followed by a ReLU function, having the same size of the input vectors and the second one of size 128. $\tau$ is set 0.07. The batch size is 256. We use stochastic gradient descent (SGD) for opimization. The initial learning rate is 0.05. We add contrative learning on the representations from the last two residual blocks. 

When pre-train the encoder for FSL problem, in addition to the multi-level contrastive leanring, we also adopt supervised learning. We let the encoder followed by a classifier make classifications for the training dataset. 
The total loss in pre-training phase is the sum of cross-entropy loss from supervised learning and mutli-level self-supervised contrastive loss.

After pre-training, we add shift and scaling parameters for the convolutional layers in the encoder and train the parameters by the MTL approach used in~\citep{sun2019meta}. To further improve generalization, we also fine-tune the upper layers in an encoder using $D$\textsubscript{val} as unlabelled data, following the method proposed in~\citep{han2020automatically}. Specifically, we fine-tune the upper half layers of the encoder while freezing the rest. After fine-tuning, the encoder remains unchanged. 
If multiple encoders are employed, each of them will undergo the process independently. 

The evaluation of the ensembles was conducted using $C$-way $N$-shot classification tasks with $C=5$ and $N=1$ or $N=5$. In an episode of evaluation, a classification task was constructed by randomly selecting $C$ classes and $N$ samples per class from $D_{test}$ to serve as the support set. 15 random query data points from the $C$ classes were also selected as the query images to test the classification. The evaluation process consisted of 1000 episodes. 
The mean accuracy (in \%) over the 1000 episodes 
and the 95\% confidence interval are reported in the experiment results. 
Input images were re-scaled to the size $80\times80$ and normalized before fed into the model. The relation networks to produce the similarity scores between the support and the query images were composed of 2 convolutional layers followed by a fully-connected layer with a sigmoid function. 
Our implementation is based on PyTorch (Paszke et al., 2017a). Pre-training phase use default parameters in the works we cited. For optimization of the relation networks , we use SGD with the Nesterov momentum 0.9. The initial learning rate is set to 1e-2.

\subsection{Analysis of Multi-Level Contrastive Learning}
\label{sec:cl_adv}
\begin{figure*}
\centering
\includegraphics[scale=1.0]{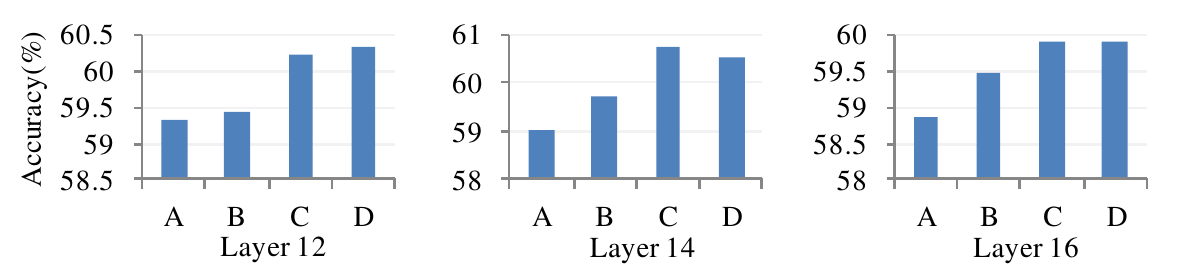}
\caption{Few-shot classification accuracy (\%) of representations from layer 12, 14 and 16 of ResNet-18 under four contrastive learning settings: (A) no contrastive learning, (B) contrastive learning on layer 16, (C) contrastive learning on layer 12 and 16, and (D) contrastive learning on layer 12, 14 and 16.}
\label{fig:CL}
\end{figure*}

We performed a number of experiments to investigate the first question we asked in the introduction: can multi-level contrastive learning lead to better representations for downstream tasks?
We chose the 5-way 1-shot classification problem in mini-ImageNet dataset as an example. We trained the encoder (ResNet-18) under 4 settings: only supervised learning without contrastive learning ((A) in Figure \ref{fig:CL}), supervised learning with contrastive learning on the last layer of the encoder (layer 16 )(B), contrastive learning on layer 12 and 16 (C), and on layer 12, 14 and 16 (D). After pre-training the encoder, we extracted the representations from layer 12, 14 and 16, and tested their classification performance on testing dataset of mini-ImageNet. 


\textbf{Multi-level CL boosts the performance of representation from the top layer.} The rightmost of Figure \ref{fig:CL} shows the classification results of layer 16 (the top layer) in ResNet-18 under 4 contrastive learning settings. Apparently, the worst accuracy comes from the scenario (A) when contrastive learning is not employed at all. The best two results are from the settings (C) and (D), where contrastive learning is adopted in the layer 16 as well as the intermediate layers (layer 12 and layer 14). Compared with the setting (B) when only adding contrastive loss on the layer 16 itself, involving more contrastive loss on the intermediate layers improves the classification performance of the layer 16 from 59.5\% to around 59.9\%.

\textbf{Multi-level CL enhances the performance for intermediate layers}. The leftmost and the middle of Figure \ref{fig:CL} show the classification accuracy of two intermediate layers: layer 12 and 14. For both of them, the lowest accuracy is from setting (A). The best accuracy for layer 12 (60.4\%) is produced in (D) when contrastive loss is applied on layers 12, 14 and 16. However the the top accuracy for layer 14 (60.7\%) comes from (C) where layers 12 and 16 have contrastive loss.
The classification accuracy of layer 12 is nearly 59.5\% in setting (B), and the accuracy increases into 60.2\% in (C). Similar improvement is also obtained for layer 14, accuracy improved from 59.7\% in setting (B) to 60.7\% in setting (C). Thus these two layers confirm that using multi-level contrastive learning (setting C and D) yields better performance than single-level counterpart (setting B).
Meanwhile we notice that the increase of accuracy (0.7--~0.9\%) generated by using multi-level contrastive learning for intermediate layers are bigger than the improvement (0.4\%) of layer 16.

\textbf{Intermediate layer selection on multi-level CL}. It seems that adding contrastive loss on more intermediate layers provides bigger benefit for the representation performance. But we observe from Figure \ref{fig:CL} that (1) the best performance for layer 14 comes from setting (C) rather than (D), which means aggregating more layers into multi-level contrastive learning does not profit layer 14; and (2) the improvement of adding more contrastive loss is trivial under certain circumstances. For example, the accuracy of layer 16 in setting (D) is 59.91\% and that in (C) is 59.89\%. Layer 12 behaves likewise, 60.2\% in setting (D) and 60.3\% in (C). We reason that the layer 14 and 16 are from the same residual block, so their representations share the most features, which may result in the insignificance of setting (D). This inspired us to apply contrastive learning only on the final layer of each residual block in ResNet encoder.

Through above analysis, we show that multi-level contrastive learning lead to better representation. There is one more evidence in Figure \ref{fig:CL} that suggests the necessity of using multi-level contrastive learning if we are interested in the representations from intermediate layers.
The enhancement of accuracy from setting (A) to (B) shows the efficiency of applying a single-level contrastive learning. The classification results of layer 14 and 16 in setting (A) and (B) show that the contrastive learning on layer 16 is significantly beneficial for these two layers. The accuracy is increased by 0.7\% and 0.6\% separately (the middle and rightmost of Figure \ref{fig:CL}). But the benefit from single-level contrastive learning is minor for the lower-level layer, the layer 12, accuracy increased by 0.1\% only (the leftmost of Figure \ref{fig:CL}). By adding contrastive learning on the layer 12, its performance is then boosted largely.

\subsection{Combining Multi-Level CL with Multi-Representation Ensemble for FSL}
We build an ensemble utilizing the multi-level representations in the FSL problems. The ensemble used in this section employed representations from the last (from top) 9 convolutional layers in ResNet-18 and the last (from top) 6 layers in ResNet-18-v1. The results on the two benchmark datasets are shown
in Table~\ref{tab:main-re}. 
Since we do not make use of extra data points, we only compare our results to the best prior results on traditional FSL.
Table~\ref{tab:main-re} shows that our model that combines multi-level contrastive learning with an ensemble gives the new state-of-the-art performance on the 1-shot and the 5-shot tasks for both the mini-ImageNet and the tiered-ImageNet datasets. Meanwhile the comparison of the last two records in Table~\ref{tab:main-re}, and the last two columns in Table~\ref{tab:multi-level}, on average 2--3\% performance gain, illustrates the effectiveness of multi-level contrastive learning for the FSL problems.

\begin{table}[t]
\caption{ The 5-way, 1-shot and 5-shot classification accuracy (\%) on mini-ImageNet and tiered-ImageNet datasets. Average classification performance over 1000 randomly generated episodes, with 95\% confidence intervals. Ours(NoCL) means our proposed method but without contrastive learning. (* Confidence intervals are not reported in the original paper.)}
\label{tab:main-re}
\begin{tabular}{llllll}
\hline 
\multirow{2}{*}{\bf Model} & \multirow{2}{*}{\bf Encoder} &\multicolumn{2}{c}{\bf mini-ImageNet} &\multicolumn{2}{c}{\bf tiered-ImageNet} \\ 
\multicolumn{1}{c}{} &\multicolumn{1}{c}{} &1-shot & 5-shot &1-shot & 5-shot\\ 
\hline 
TADAM \citep{oreshkin2018tadam} &ResNet-12 &$58.50\pm0.30$ &$76.70\pm0.30$ &- &- \\
MTL \citep{sun2019meta} &ResNet-12 &$62.10\pm1.80$ &$78.50\pm0.90$ &$67.8\pm1.8$ &$83.0\pm0.7$ \\
TapNet \citep{yoon2019tapnet} &ResNet-12 &$61.65\pm0.15$ &$76.36\pm0.10$ &$63.08\pm0.15$ &$80.26\pm0.12$ \\
MetaOpt-SVM \citep{lee2019meta} &ResNet-12 &$62.64\pm0.61$ &$78.63\pm0.46$ &$65.99\pm0.72$ &$81.56\pm0.53$ \\
CAN \citep{hou2019cross} &ResNet-12 &$63.85\pm0.48$ &$79.44\pm0.34$ &$69.89\pm0.51$ &$84.23\pm0.37$ \\
CTM \citep{li2019finding} &ResNet-18 &$64.12\pm0.82$ &$80.51\pm0.13$ &$68.41\pm0.39$ &$84.28\pm1.73$ \\
FEAT \citep{ye2020few} &ResNet-18 &66.78 &82.05 &$70.80\pm0.23$ &$84.79\pm0.16$ \\
\hline
\bf Single-level CL \\
S-MoCo \citep{majumder2021revisiting} &ResNet-18 &$59.94\pm0.89$ &$78.17\pm0.64$ &$68.70\pm0.72$ &$84.40\pm0.64$ \\
SC \citep{ouali2020spatial} &ResNet-18 &$67.40\pm0.76$ &$83.19\pm0.54$ &$71.98\pm0.91$ &$86.19\pm0.59$ \\
CPLAE \citep{gao2021contrastive} &ResNet-12 &$67.46\pm0.44$ &$83.22\pm0.2$ &$72.23\pm0.50$ &$87.35\pm0.34$ \\
\hline
\bf Ensemble \\
Robust-dist++  \citep{dvornik2019diversity} &ResNet-18 & $59.48\pm0.62$ &$75.62\pm0.48$ &- &- \\
MTL+E$^{3}$TB* \citep{liu2019ensemble} &ResNet-25 &64.3 &81.0 &70.0 &85.0 \\
Ours(NoCL) &ResNet-18 &65.01$\pm$0.35  &82.31$\pm$0.36 &70.87$\pm$0.37 &85.31$\pm$0.42 \\
\hline
\bf MLCL + Ensemble \\
Ours &ResNet-18 & \ubold{68.96$\pm$0.66}  &\ubold{84.51$\pm$0.49} &\ubold{73.49$\pm$0.57} &\ubold{88.72$\pm$0.31} \\
\hline 
\end{tabular}
\end{table}


\subsection{Benefit of Lower-Level Representations}

In this section, we further demonstrate the benefit of using lower-level features 
and discuss the design choices for constructing a better ensemble. 
To investigate the ensemble strategy design, we first present and discuss the performance results from ensembles that involve multi-representation from a single encoder and then continue to ensembles that employ both multiple representations and encoders. The ensemble performances (accuracy) were measured for few-shot classification on the mini-ImageNet dataset. 

\textbf{Single Encoder Multiple Representation}.
The best accuracy obtained by a classifier using a single representation from ResNet-18 is 60.81\% (with representation from layer 13). In Table~\ref{tab:multi-level}, we observe that ensembles utilizing multi-representation have boosted performance. For example, the ensemble containing classifiers using representations from layer 16 to layer 14 of ResNet-18 achieves an accuracy of 65.19\% (row 1 of Table~\ref{tab:multi-level}). Including more classifiers that use representations from layers 13, 12 and 11, the ensemble can reach an even higher performance (66.04\%, row 2). The topmost accuracy (66.84\%) from the ensembles based on ResNet-18 comes from the one that utilizes representations from layers 16 to 8. The improvement by including multiple representations into the ensemble can also be observed for the ResNet-18-v1 and the ResNet-18-v2 encoders. In general, for single encoder ensembles, coming down from the top layer, the more representations we include in the ensemble, the better performance the ensemble can achieve, until a turning-point layer is reached. Adding representations after the turn point may lead to reduced performance.

\textbf{Multiple Encoder Multiple Representation}.
Incorporating different models (e.g., neural networks of different structures) is a common method to construct an ensemble. Clearly, this multi-model construction can be combined with our multi-representation construction to create ensembles that employ both multiple models and multiple representations from each model. The last row of Table~\ref{tab:multi-level} shows the configuration and the performance of one of our multi-model, multi-representation ensembles used in the experiments. There is a significant performance gain comparing this ensemble to those on row 13 and 14. The ensembles on row 13 and 14 are multi-model but not multi-representation since only one representation is used from each of the encoders in the ensembles. Although a multi-model ensemble already performs better than the individual models in the ensemble, adding multi-representation on top of multi-model leads to even better performance, giving rise to the best performer among the ensembles (row 15).

\begin{table}[t]
\caption{Classification accuracy (\%) of multi-model multi-representation ensembles. The last column Accuracy(NoCL) shows the results from the model without contrastive learning. (The second column lists encoders and the third lists layers (or ranges of layer) from which representations were used. If there were multiple encoders, their layer(s) information is separated by a comma.)} 
\label{tab:multi-level}
\begin{center}
\begin{tabular}{lllll}
\multicolumn{1}{c}{\bf Row} &\multicolumn{1}{c}{\bf Encoder(s)}  &\multicolumn{1}{c}{\bf Conv layers} &\multicolumn{1}{c}{\bf Accuracy} 
&\multicolumn{1}{c}{\bf Accuracy(NoCL)}
\\ 
\hline 
1 &ResNet-18           &16--14 &65.19 &62.59\\
2 &ResNet-18           &16--11 &66.04 &63.54\\
3 &ResNet-18           &16--8 &66.84 &64.03\\
4 &ResNet-18           &16--6 &66.12 &63.71\\
5 &ResNet-18-v1        &18--17 &60.95 &58.57\\
6 &ResNet-18-v1        &18--15 &63.21 &59.67\\
7 &ResNet-18-v1        &18--12 &63.38 &60.72\\
8 &ResNet-18-v1        &18--11 &62.19 &59.21\\
9 &ResNet-18-v2        &16--14 &58.55 &56.02\\
10 &ResNet-18-v2        &16--12 &69.48 &57.49\\
11 &ResNet-18-v2        &16--10 &61.66 &58.26\\
12 &ResNet-18-v2        &16--8 &59.27 &58.71\\
13 &ResNet-18, ResNet-18-v1        &16, 18 &63.04 &60.08\\
14 &ResNet-18, ResNet-18-v1, ResNet-18-v2        &16, 18, 16 &61.91 &58.99\\
15 &ResNet-18, ResNet-18-v1        &16--8, 18--13 &68.96 &65.01\\
\hline
\end{tabular}
\end{center}
\end{table}

\section{RELATED WORK}
\textbf{Contrastive learning}. 
Contrastive learning is a discriminative approach that aims at grouping similar samples closer and diverse samples far from each other. A similarity metric is used to measure how close two embeddings are. Contrastive learning can be self-supervised\cite{chen2020simple}, semi-supervised \cite{inoue2020semi}  or supervised \cite{khosla2020supervised} depending on if or how it utilizes data labels. In self-supervised setting, embeddings and their augmentations (such as an image and its rotated counterpart) as the same class, and all others as different categories. Contrastive learning has been applied into FSL problems and achieved much progress \cite{li2020few, gao2021contrastive, majumder2021revisiting}. \citet{li2020few} employed Contrastive learning to train a transferable encoder via contrastive self-supervised learning. \citet{gao2021contrastive} proposed a supervised contrastive prototype learning, which makes use of the prototype of image embeddings. Our work applies self-supervised contrastive learing for multiple representations from the a neural network.

\textbf{Ensemble methods}. 
Ensemble methods are commonly used to improve prediction quality. Some example ensemble strategies include: (1) manipulate the data, such as data augmentation or dividing the original dataset into smaller subsets and then training a different model on each subset. 
(2) apply different models or learning algorithms. For example,  train a neural network with varied hyperparameter values such as different learning rates or different structures. 
(3) hybridize multiple ensemble strategies, e.g., random forest. 
Ensembles have also been applied to FSL problems. \citet{liu2019lcc} proposed to learn an ensemble of temporal base-learners, which are generated along the training time, producing encouraging results on the mini-ImageNet and the Fewshot-CIFAR100 datasets. \citet{dvornik2019diversity} introduced mechanisms to encourage cooperation and diversity of individual classifiers in an ensemble model. The main difference between our method and the previous ones is the ensemble construction that utilizes multiple representations from different depth of a neural network.

\textbf{Few-shot learning}. 
Meta-learning method has shown great success in FSL \citep{finn2017model,grant2018recasting, lee2018gradient}. 
MAML \citep{finn2017model} used a meta-learner that learns from the training images to effectively initialize a base-learner for a new learning task in the test dataset. Further works aimed to enhance the generalization ability by improving the learning algorithm \citep{nichol2018first}, fine-tuning the image embedding space \citep{sun2019meta, rusu2018meta}. Another popular direction for FSL is metric-learning which targets on learning metric space where classes can be easily separated. For example, Prototypical Networks use euclidean distance with each class prototype set to be the mean of the support embeddings \citep{snell2017prototypical}. Relation network \citep{sung2018learning} was proposed to compute the similarity score between a pair of support and query images. The query image is classified into the category with the highest similarity. Each individual model in our ensemble employs a relation network for classification while the relational network in different models receives different representation as input. Many recent FSL studies proposed approaches that utilized extra unlabled data or other additional information \citep{li2019learning,kye2020transductive, yang2020dpgn, hu2020exploiting, rodriguez2020embedding, lichtenstein2020tafssl}. They are not in the scope of the problem we were considering and thus not compared in the result section.

\section{CONCLUSIONS}
\label{conclusions}
In this paper we propose a multi-level contrastive learning approach, in which multiple contrastive losses are added into the representations from different layers of an encoder, aiming to increase the feature difference of these representations among classes. We test the efficiency of the proposed method in few-shot learning problems. Specifically, we build an ensemble framework that creates an ensemble of classifiers, each using the representation/feature map from a different depth in a CNN encoder.  Through experiments, we validated the effectiveness of our approach. 
Our model achieved the new state-of-the-art results on two commonly-used FSL benchmark datasets. We further conducted experiments and analysis to investigate the benefits of adding multi-level contrastive learning for the encoders and 
examined the selection of representations for creating a better ensemble. It is quite possible that the approaches we studied here (multi-level contrastive learning and multi-representation ensemble) are not limited to the few-shot learning problems. As future work, we plan to explore more scenarios where these approaches can be applied, as well as to refine the multi-level strategies employed in the approaches.     



\bibliography{iclr2021_conference.bib}
\bibliographystyle{plainnat}

\end{document}